\pdfoutput=1
\documentclass[11pt]{article}

\usepackage[final]{acl}

\usepackage{times}
\usepackage{latexsym}
\usepackage{graphicx} % Required for including graphics

\usepackage[T1]{fontenc}
\usepackage[utf8]{inputenc}

\usepackage{microtype}
\usepackage{booktabs}
\usepackage{multirow}
\usepackage{pifont}
\usepackage[most]{tcolorbox}

\title{FlexRAG: A Flexible and Comprehensive Framework for Retrieval-Augmented Generation}

\author{Zhuocheng Zhang\textsuperscript{1,2}, Yang Feng\textsuperscript{1,2,3}\footnotemark[1], Min Zhang\textsuperscript{4} \\
  \textsuperscript{1}Key Laboratory of Intelligent Information Processing, \\
  Institute of Computing Technology, Chinese Academy of Sciences (ICT/CAS) \\
  \textsuperscript{2}University of Chinese Academy of Sciences, China \\
  \textsuperscript{3}Key Laboratory of AI Safety, Chinese Academy of Sciences \\
  \textsuperscript{4}Institute of Computing and Intelligence, Harbin Institute of Technology (Shenzhen), China \\
   \texttt{\href{mailto:zhangzhuocheng20z@ict.ac.cn}{zhangzhuocheng20z},\href{mailto:fengyang@ict.ac.cn}{fengyang}@ict.ac.cn
   }
   \texttt{\href{mailto:minzhang@suda.edu.cn}{minzhang}@suda.edu.cn
   }
  }

\begin{document}
\maketitle
\renewcommand{\thefootnote}{\fnsymbol{footnote}}
\footnotetext[1]{Corresponding author.}
\renewcommand{\thefootnote}{\arabic{footnote}}
\begin{abstract}
Retrieval-Augmented Generation (RAG) plays a pivotal role in modern large language model applications, with numerous existing frameworks offering a wide range of functionalities to facilitate the development of RAG systems.
However, we have identified several persistent challenges in these frameworks, including difficulties in algorithm reproduction and sharing, lack of new techniques, and high system overhead.
To address these limitations, we introduce \textbf{FlexRAG}, an open-source framework specifically designed for research and prototyping.
FlexRAG supports text-based, multimodal, and network-based RAG, providing comprehensive lifecycle support alongside efficient asynchronous processing and persistent caching capabilities.
By offering a robust and flexible solution, FlexRAG enables researchers to rapidly develop, deploy, and share advanced RAG systems.
Our toolkit and resources are available at \href{https://github.com/ictnlp/FlexRAG}{https://github.com/ictnlp/FlexRAG}.
\end{abstract}

\section{Introduction}

With the rapid advancement of large language models (LLMs) \cite{GPT4,LLama3,Qwen2}, they are increasingly playing a pivotal role across various domains. However, numerous application scenarios necessitate that these models maintain accurate, comprehensive, and up-to-date knowledge \cite{SurveyRAG1,SurveyRAG2}. Continuously retraining LLMs to integrate new information is not only computationally expensive but also poses challenges such as catastrophic forgetting. To address these limitations, retrieval-augmented generation (RAG) has emerged as a promising solution, enabling models to dynamically retrieve relevant information from external sources, thereby enhancing their factual accuracy and adaptability.

Given the vast application potential of RAG across various domains, numerous frameworks have emerged in recent years to facilitate rapid construction of RAG systems \cite{FlashRAG,Ralle,EasyRAG,RAGLab,AutoRAG,LocalRQA}.
However, a comprehensive analysis of existing frameworks reveals that these tools still fail to adequately address several core challenges in RAG research.
% First, the lack of standardized retrievers and evaluation workflow significantly hinders the reproducibility and comparative analysis of different approaches, posing challenges for the broader research community.
% Second, current frameworks often suffer from excessive memory consumption and high computational costs, which impose substantial resource constraints on researchers and limit the feasibility of large-scale experimentation.
% Most critically, existing frameworks frequently overlook the practical requirements of real-world applications, leading to a persistent gap between academic research and industrial adoption.
% These challenges not only impede the effective deployment of RAG technology in production environments but also constrain its further advancement and innovation.
First, due to the complexity of RAG systems, which involve multiple components and intricate environment configurations, researchers often struggle to precisely reproduce existing studies or effectively share their own work with others.
Second, constructing a RAG system is inherently complex, requiring researchers to address numerous engineering challenges, which significantly diverts their focus from scientific inquiry.
Furthermore, as RAG technology evolves rapidly, many researchers are exploring advanced topics such as multimodal retrieval, web-based retrieval, and document chunking.
However, most existing frameworks are designed to address only a single aspect of RAG research, specifically retrieval strategies.
More importantly, both the retrieval and generation components in RAG systems impose substantial computational costs, limiting the ability of resource-constrained researchers to conduct effective investigations.

To address these issues, we present FlexRAG, a novel open-source framework designed to facilitate the rapid reproduction, development, and evaluation of RAG systems.
The proposed framework offers comprehensive support for diverse RAG scenarios, including text-based, multimodal, and web-accessible RAG applications, while providing end-to-end pipeline support from data preparation to system evaluation.
FlexRAG enables researchers to efficiently share their work with the community and quickly develop demonstrative prototypes based on their algorithms.
Notably, FlexRAG incorporates four key distinguishing features that set it apart from existing frameworks, which are as follows.

\begin{figure*}[ht]
  \centering
  \includegraphics[width=\linewidth]{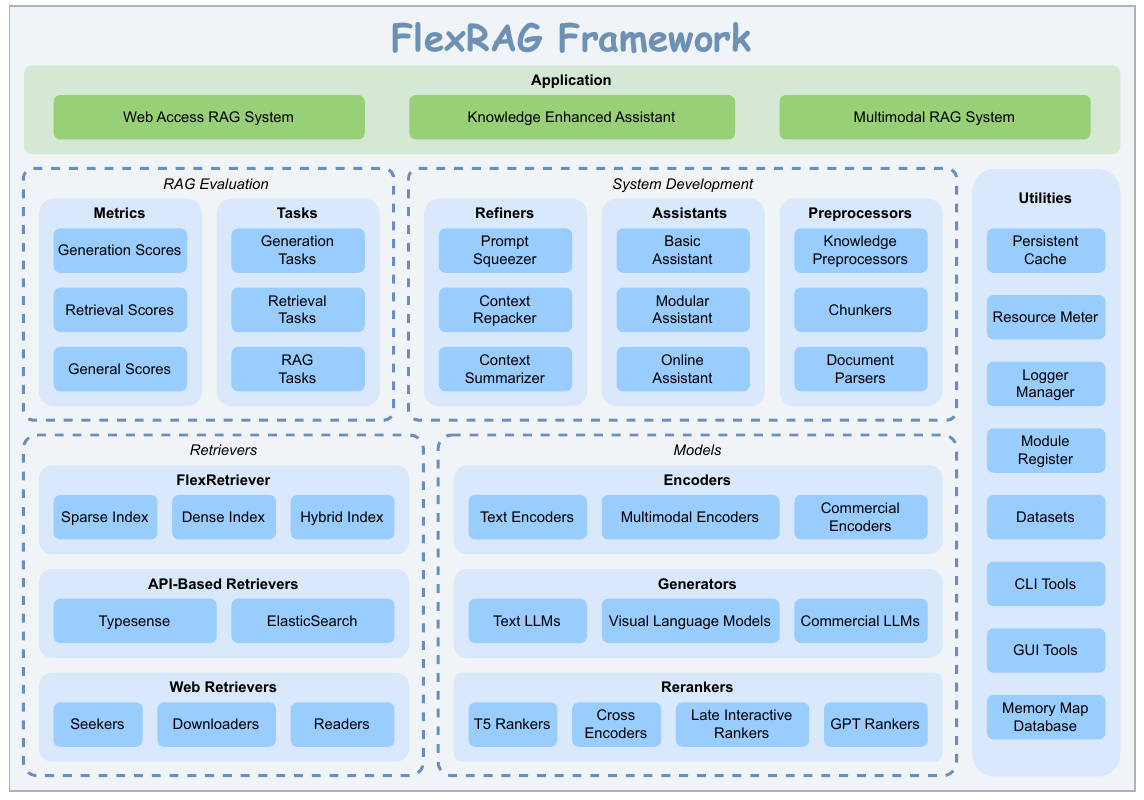}
  \caption{The architecture of FlexRAG. The light blue boxes represent modules, while the dashed boxes indicate collections of modules with relevant functions.}
  \label{fig:architecture}
\end{figure*}

\paragraph{Research Oriented Design}
% FlexRAG standardizes retriever implementation, enabling reproducible system development in a unified environment.
% By integrating with Hugging Face Hub, FlexRAG facilitates community-driven retriever sharing, enhancing collaborative research.
% The framework also provides benchmark implementations of RAG algorithms and a standardized evaluation protocol, ensuring fair and efficient performance comparisons.
FlexRAG provides a unified configuration management system and a standardized RAG evaluation process to ensure fair and convenient performance assessment.
By integrating with the Hugging Face Hub, FlexRAG enables researchers to share their retrievers with the community, fostering collaborative research efforts.
Moreover, FlexRAG offers an example repository that facilitates algorithm comparison and reproduction, supporting rigorous scientific inquiry.

% \paragraph{Low Learning Curve}
% The framework allows users to effortlessly load pre-built retrievers with minimal setup requirements.
% Additionally, it standardizes configuration management across RAG components while providing optimized default parameters that balance retrieval accuracy with computational efficiency.
% Its modular architecture offers exceptional flexibility, accommodating a wide range of use cases.
% To further support developers, FlexRAG includes a comprehensive bilingual (English-Chinese) documentation for seamless system implementation.

\begin{figure}[t]
  \centering
  \begin{tcolorbox}[enhanced,
    enhanced,
    colframe=white,
    colback=white,
    boxrule=0pt,
    sharp corners,
    drop shadow={black!40!white},
    left=0pt,
    right=0pt,
    top=0pt,
    bottom=0pt,
    boxsep=0pt,
    halign=center,
  ]
    \includegraphics[width=\linewidth]{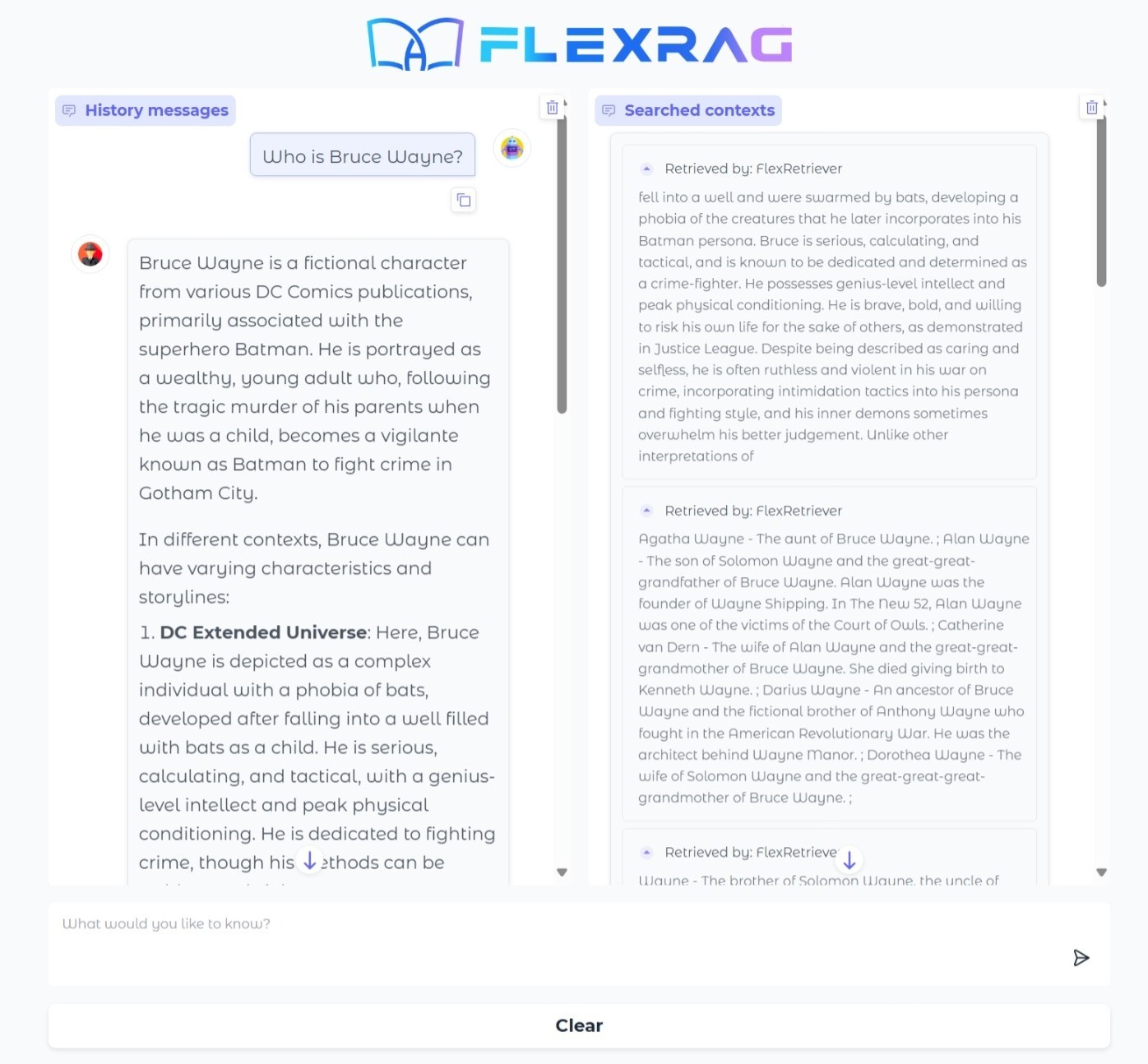}
  \end{tcolorbox}
  \caption{The GUI demonstration provided by FlexRAG. The left panel displays the messages exchanged between the user and the assistant, while the right panel shows the retrieved contexts. The GUI is designed to facilitate user interaction with the RAG system, allowing users to input queries and receive responses in a user-friendly manner.}
  \label{fig:gui}
\end{figure}

\paragraph{Extensive Infrastructure and Tooling} To reduce the engineering burden on researchers, FlexRAG provides complete bilingual documentation and pre-built retrievers available on the Hugging Face Hub, facilitating the rapid implementation of algorithms.
Additionally, FlexRAG provides a comprehensive command-line toolkit that facilitates a wide range of tasks, including data preprocessing, retriever construction, system evaluation, and the development of GUI prototypes, as illustrated in Figure~\ref{fig:gui}.

% \paragraph{Diverse Application Scenarios}
% FlexRAG supports not only text-based RAG but also multimodal and network-based RAG, enabling broad applications across different data types. 
% The framework provides end-to-end support for the entire RAG pipeline, including document parsing, chunking, and more, empowering the development of comprehensive RAG systems.
% Notably, FlexRAG adopts the permissive MIT open-source license, encouraging widespread adoption and community contributions.

\paragraph{Comprehensive Technical Support}
FlexRAG not only supports text-based RAG but also extends to multimodal and web-based RAG, enabling broad applicability across various data types.
Additionally, the framework provides end-to-end support for the entire RAG pipeline, including document parsing, chunking, and other essential processes, facilitating the development of comprehensive RAG systems.

\paragraph{Superior Performance}
% By offering asynchronous functions for computationally intensive components, FlexRAG enables the development of high-throughput RAG system prototypes.
% In addition, it employs a persistent caching mechanism that further reduces retrieval overhead and improves retrieval efficiency.
% Collectively, these features enhance the overall performance and scalability of the system, establishing FlexRAG as a robust solution for resource-intensive retrieval tasks.
% FlexRAG adopts a modular architecture design and integrates asynchronous functions for computationally intensive components, facilitating the development of high-throughput RAG system prototypes.
FlexRAG employs a modular design and leverages asynchronous functions for computationally intensive components, facilitating the development of high-throughput RAG system prototypes.
Moreover, it employs a persistent caching mechanism to further reduce retrieval overhead and enhance retrieval efficiency.
Most importantly, FlexRAG incorporates advanced indexing techniques and memory map mechanism, consuming only one-tenth of the CPU and memory resources required by comparable frameworks when performing large-scale retrieval tasks.

In summary, FlexRAG is a comprehensive and flexible framework that addresses the core challenges of RAG research, providing researchers with a powerful tool to develop, evaluate, and deploy RAG systems.

% \begin{figure}[t]
%   \centering
%   \includegraphics[width=\linewidth]{./pics/FlexRAG_GUI.jpeg}
%   \caption{The GUI demonstration provided by FlexRAG. The left panel displays the messages exchanged between the user and the assistant, while the right panel shows the retrieved contexts. The GUI is designed to facilitate user interaction with the RAG system, allowing users to input queries and receive responses in a user-friendly manner.}
%   \label{fig:gui}
% \end{figure}

\section{The Architecture of FlexRAG}

As illustrated in Figure~\ref{fig:architecture}, FlexRAG comprises twelve core modules, each serving a distinct function in the RAG pipeline.
For clarity, we categorize them into four functional groups: models, retrievers, system development, and evaluation, along with auxiliary utility tools.
This section first introduces the modules within these four categories, followed by a detailed discussion of the remaining components.

\subsection{Models}
In RAG systems, models are employed across various components. For instance, dense retrievers utilize encoders to transform knowledge pieces into dense vector representations, while generators are required to produce final responses. FlexRAG incorporates three fundamental model categories: encoders, generators, and rerankers.

\paragraph{Encoders} The encoder functions to convert input queries or documents into dense vectors for similarity search in vector space. The encoders in FlexRAG can be classified into text encoders \cite{BERT,Contriever,DPR,Dragon} and multimodal encoders \cite{CLIP} based on input data types. Additional, FlexRAG also supports calling commercial encoders via API calls\footnotemark[1]\textsuperscript{,}\footnotemark[2]\textsuperscript{,}\footnotemark[3], and deploying encoders using famous frameworks\footnotemark[4]\textsuperscript{,}\footnotemark[5].

\footnotetext[1]{https://jina.ai/}
\footnotetext[2]{https://cohere.com/}
\footnotetext[3]{https://www.openai.com/}
\footnotetext[4]{https://ollama.com/}
\footnotetext[5]{https://sbert.net/}

\paragraph{Rerankers} Rerankers optimize the initially retrieved document list through reordering mechanisms, effectively filtering out irrelevant content to reduce noise and enhance input quality for generators. FlexRAG supports various rerankers, including cross-encoder rerankers\cite{BGE_M3}, late-interaction rerankers\cite{ColBERT,ColBERTv2,JinaColBERT}, T5-style rerankers\cite{RankT5}, and GPT-style rerankers\cite{RankLLM}. In addition, FlexRAG also supports calling online rerankers via APIs\footnotemark[1]\textsuperscript{,}\footnotemark[2]\textsuperscript{,}\footnotemark[6]\textsuperscript{,}\footnotemark[7].

\footnotetext[6]{https://www.mixedbread.com/}
\footnotetext[7]{https://www.voyageai.com/}

\paragraph{Generators} The generator synthesizes natural language responses based on the retrieved documents and user queries. FlexRAG implement traditional LLMs \cite{Qwen2,LLama3} and Vision Language Models (VLMs) \cite{Qwen2_VL,PaliGemma2} to serve as generators.
Similarly, FlexRAG supports calling commercial generators via API calls\footnotemark[3]\textsuperscript{,}\footnotemark[8], or deploying generators using fast inference engines\footnotemark[4]\textsuperscript{,}\footnotemark[9].

\footnotetext[8]{https://www.anthropic.com/}
\footnotetext[9]{https://github.com/vllm-project/vllm}

\begin{figure}[t]
  \centering
  \includegraphics[width=\linewidth]{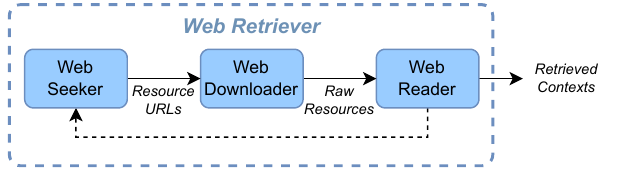}
  \caption{The core components of the \emph{WebRetriever} module in FlexRAG and its typical workflow.}
  \label{fig:web_retriever}
\end{figure}

\subsection{Retrievers}
The retriever constitutes one of the most critical components in RAG systems, serving to rapidly identify relevant information based on user queries.
% In FlexRAG, retrievers can be systematically categorized through a hierarchical classification process: primarily differentiated as web retrievers versus knowledge base retrievers based on their data provenance, followed by subsequent division of knowledge base retrievers into local retrievers and remote retrievers according to the storage locality of the knowledge base.
In FlexRAG, retrievers are classified into three types: the \textit{Web Retriever}, which gathers information directly from the internet; the \textit{API-Based Retriever}, which connects to external retrieval systems via APIs; and the \textit{FlexRetriever}, developed in-house by the FlexRAG team, which stores the knowledge base locally and builds indexes using sparse, dense, or hybrid techniques.

\subsubsection{Web Retrievers}
Web retrievers are designed to retrieve information from the internet, typically through search engines or walking through web pages.
With internet access, web retrievers has significant advantages in both the timeliness of retrieval and the breadth of information it can access, making them particularly suitable for building personal assistants.

As shown in Figure~\ref{fig:web_retriever}, FlexRAG designs three key roles to support the construction of web retrievers.
The \emph{Web Seeker} is responsible for locating online resources. It can be implemented as either a search engine interface or a web crawler.
The \emph{Web Downloader} handles the downloading of web resources.
Since web resources are typically in HTML format, which is challenging for LLMs to process directly, the \emph{Web Reader} is designed to extract content from raw web pages.

To further streamline the development process of RAG systems, FlexRAG provides two built-in web retrievers: the \emph{SimpleWebRetriever}, which leverages search engines to locate web pages and employs a Web Reader to convert them into an LLM-friendly format, and the \emph{WikipediaRetriever}, specifically designed for direct entity querying from Wikipedia knowledge bases.

\subsubsection{FlexRetriever}
% Local retrievers, which are designed to extract relevant information from local knowledge bases, play a crucial role in scientific research.
% FlexRAG implements two types of local retrievers: a sparse retriever based on the BM25 algorithm \cite{BM25S} and a dense retriever leveraging vector indexing techniques \cite{Faiss,ScaNN}.
% Notably, FlexRAG supports a variety of vector indexing algorithms, allowing researchers to optimize indexing configurations based on hardware constraints and the scale of the knowledge base.
% In particular, FlexRAG employs Inverted File and Product Quantization (IVFPQ) indexing techniques with carefully optimized parameters \cite{ANNBench} as its default configuration, achieving significantly lower memory overhead and superior retrieval efficiency compared to alternative frameworks.
FlexRetriever is a versatile retriever that supports both MultiField and MultiIndex retrieval paradigms.
It enables documents to be decomposed into multiple semantic fields, such as title, abstract, and content, with dedicated indexes constructed for each field.
Moreover, FlexRetriever facilitates hybrid retrieval across multiple indexes, allowing for flexible and fine-grained retrieval strategies that can be tailored to address complex information needs.
The system supports both sparse and dense retrieval approaches \cite{BM25S, Faiss,ScaNN}, making it applicable to a wide spectrum of retrieval tasks.

Notably, FlexRetriever employs memory map and the empirical formula \cite{ANNBench} designed for Inverted File and Product Quantization (IVFPQ) indexing techniques as its default configuration, achieving significantly lower memory overhead and superior retrieval efficiency compared to alternative frameworks.

Furthermore, FlexRetriever is fully integrated with the Hugging Face ecosystem, enabling seamless publication, sharing, and reuse of retrievers via the Hugging Face Hub \footnote[10]{https://huggingface.co/FlexRAG}.
This integration promotes community collaboration and lowers the barrier to leveraging and contributing retrieval pipelines with minimal configuration overhead.

% Moreover, all FlexRetriever is seamlessly integrated with the Hugging Face Hub, facilitating efficient sharing and reproducibility of research implementations within the academic community.

\subsubsection{API-Based Retriever}
FlexRAG also supports two API-Based Retrievers, namely TypesenseRetriever\footnote[11]{https://github.com/typesense/typesense}, and ElasticSearchRetriever\footnote[12]{https://github.com/elastic/elasticsearch}, enabling users to implement their RAG systems by leveraging mature and feature-rich retrieval systems.

\subsection{System Development}
Beyond the two fundamental modules of a RAG system, namely the retriever and the model, additional components are essential for constructing a complete RAG pipeline. To address this requirement, FlexRAG introduces three modules that collectively enhance the pipeline's functionality. 
The \textbf{Preprocessors} module is responsible for preparing and structuring the knowledge base, ensuring that relevant information is efficiently organized for retrieval.
The \textbf{Refiners} module enhances the retrieved contexts through refinement and post-processing, improving the quality and relevance of the input provided to the model.
Lastly, the \textbf{Assistants} module serves as a unified framework that encapsulates the entire RAG pipeline, facilitating seamless integration and operation.

\begin{figure}[t]
  \centering
  \includegraphics[width=1.0\linewidth]{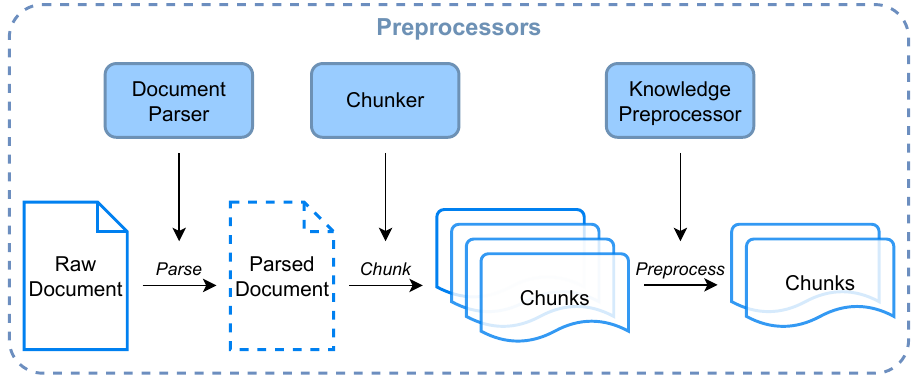}
  \caption{Architecture of the \emph{Preprocessors} module in FlexRAG and its typical workflow.}
  \label{fig:preprocessors}
\end{figure}

\begin{figure*}[t]
  \centering
  \includegraphics[width=1.0\linewidth]{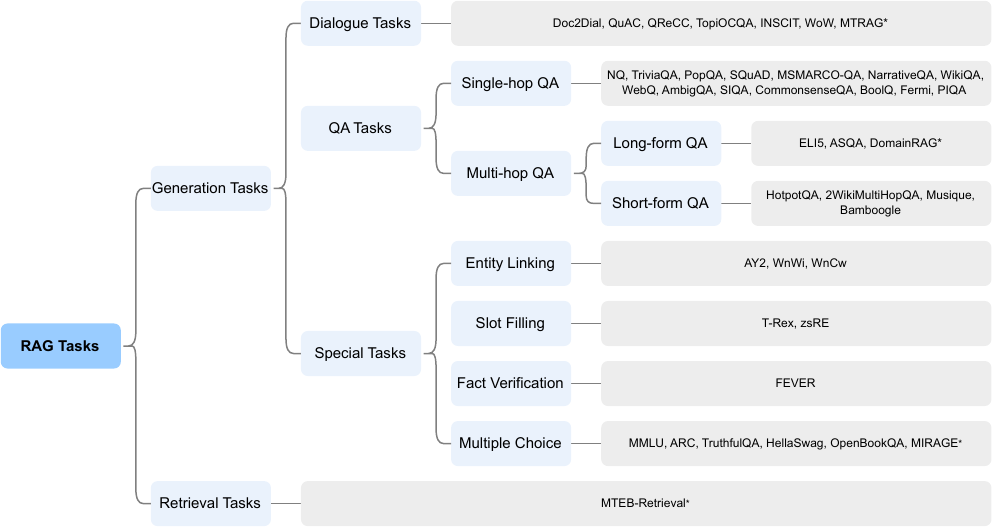}
  \caption{RAG Evaluation Tasks. Tasks without an asterisk (*) correspond to individual datasets, while those marked with an asterisk indicate benchmarks that may comprise multiple datasets.}
  \label{fig:tasks}
\end{figure*}

\subsubsection{Preprocessors}
In modern computing systems, a substantial proportion of knowledge resources are stored and disseminated through document file formats (e.g., PDF, DOCX), as opposed to plain text.
While these semi-structured data maintains human interpretability, it present significant parsing challenges for LLMs during information extraction.
To address this limitation, document preprocessing pipelines are required to transform these heterogeneous formats into standardized structured representations that are computationally tractable for LLM processing.
As illustrated in Figure~\ref{fig:preprocessors}, FlexRAG's preprocessing module comprises three specialized roles, namely \emph{Document Parser}, \emph{Chunker}, and \emph{Knowledge Preprocessor}, to facilitate this critical format conversion.

Concretely, the \emph{Document Parser} is responsible for extracting LLM readable content from various document formats, including PDF, DOCX, and HTML.
Once the content is extracted, the \emph{Chunker} segments it into smaller, more manageable chunks, enabling efficient processing by both the retriever and the generator.
To further improve knowledge base quality, the \emph{Knowledge Preprocessor} is designed to preprocess and filter the extracted content, ensuring that it is well-structured and optimized for retrieval.

\subsubsection{Refiners}
Existing research indicates that the quality of responses generated by LLMs is closely associated with the relevance, sequencing, and quantity of contextual information provided in the prompt \cite{NoisyContext,R4}. However, relying solely on retrievers does not guarantee that the retrieved context aligns with the preferences of LLMs. Therefore, further processing of the retrieved context is a critical step in constructing a high-performance RAG system. To address this issue, FlexRAG incorporates three specialized submodules: \emph{Prompt Squeezer}, \emph{Context Repacker}, and \emph{Context Summarizer}.

Concretely, the \emph{Prompt Squeezer} is designed to optimize the prompt provided to the generator, ensuring that it is concise and relevant to the user query \cite{LLMLingua,LLMLingua2,LongLLMLingua}.
To prevent critical information from being overlooked by the LLM, the \emph{Context Repacker} reorganizes the retrieved context for better coherence.
Additionally, the \emph{Context Summarizer} enhances the quality of the retrieved context by condensing it into a more concise and informative format \cite{Recomp,SuRE,CompressContext}, thereby decreasing the inference overhead.

\begin{table*}[t]
  \resizebox{\textwidth}{!}{%
\begin{tabular}{@{}lllllllllllll@{}}
\toprule
\multicolumn{1}{c}{\multirow{2}{*}{Methods}} & \multicolumn{3}{c}{PopQA(\%)} & \multicolumn{3}{c}{NQ(\%)} & \multicolumn{3}{c}{TriviaQA(\%)} & \multicolumn{3}{c}{Average(\%)} \\
\multicolumn{1}{c}{}                         & F1       & EM       & Succ    & F1      & EM      & Succ   & F1        & EM        & Succ     & F1      & EM      & Succ    \\
\cmidrule(lr){1-1} \cmidrule(lr){2-4} \cmidrule(lr){5-7} \cmidrule(lr){8-10} \cmidrule(lr){11-13}
BM25s\cite{BM25S}                            & 57.88    & 52.75    & 68.48   & 38.79   & 30.00   & 54.74  & 65.93     & 58.02     & 61.98    & 54.20   & 46.92   & 61.73   \\
Contriever*\cite{Contriever}                 & 64.14    & 59.04    & 80.77   & 49.67   & 39.03   & 75.65  & 70.36     & 62.55     & 68.26    & 61.39   & 53.54   & 74.89   \\
E5 base\cite{E5}                             & 59.74    & 54.25    & 77.20   & 50.05   & 39.56   & 78.84  & 71.66     & 63.79     & 70.63    & 60.48   & 52.53   & 75.56   \\
BGE M3\cite{BGE_M3}                          & 63.65    & 58.76    & 83.42   & 50.98   & 40.36   & 80.00  & 71.92     & 63.85     & 71.10    & 62.18   & 54.32   & 78.17   \\
\midrule
FLAT                                         & 63.65    & 58.40    & 82.20   & 49.20   & 39.11   & 77.95  & 70.61     & 62.70     & 80.03    & 61.15   & 53.40   & 80.06   \\
Faiss*\cite{Faiss}                           & 64.14    & 59.04    & 81.42   & 49.62   & 39.11   & 77.87  & 70.48     & 62.57     & 79.80    & 61.41   & 53.57   & 79.70   \\
ScaNN\cite{ScaNN}                            & 63.26    & 58.11    & 82.13   & 49.31   & 39.25   & 77.76  & 70.50     & 62.64     & 79.93    & 61.02   & 53.33   & 79.94   \\
\midrule
BGE-reranker-M3\cite{BGE_M3}                 & 66.02    & 60.76    & 86.92   & 50.94   & 40.53   & 81.91  & 74.58     & 66.71     & 84.81    & 63.85   & 56.00   & 84.55   \\
colbert-v2\cite{ColBERTv2}                   & 65.44    & 60.47    & 83.56   & 47.18   & 37.06   & 77.53  & 72.13     & 64.24     & 81.47    & 61.58   & 53.92   & 80.85   \\
InRanker-base\cite{InRanker}                 & 66.05    & 60.90    & 86.63   & 48.77   & 38.50   & 79.78  & 73.38     & 65.47     & 83.20    & 62.73   & 54.96   & 83.20   \\
rankGPT\cite{RankLLM}                        & 63.11    & 58.26    & 77.91   & 49.50   & 39.06   & 75.90  & 70.13     & 62.31     & 79.11    & 60.91   & 53.21   & 77.64   \\
\midrule
Qwen2-7B*\cite{Qwen2}                        & 64.14    & 59.04    & 81.42   & 49.62   & 39.11   & 77.87  & 70.48     & 62.57     & 79.80    & 61.41   & 53.57   & 79.70   \\
Llama3.1-8B\cite{LLama3}                     & 63.20    & 55.83    & 81.42   & 47.58   & 35.73   & 77.87  & 71.75     & 62.97     & 79.80    & 60.84   & 51.51   & 79.70   \\
ChatQA2-7B\cite{ChatQA2}                     & 60.36    & 53.82    & 81.42   & 49.84   & 39.09   & 77.87  & 71.84     & 62.67     & 79.80    & 60.68   & 51.86   & 79.70   \\
\bottomrule
\end{tabular}%
}
\caption{The experimental results of the \emph{ModularAssistant} on three widely used RAG tasks. The experiment was divided into four groups, each investigating the impact of modifying the retriever, index, re-ranker, and generator on the overall RAG system. Items marked with an asterisk in the table indicate the default configuration for this experiment. We did not use rerankers except in the experiments investigating the differences between them.}
\label{tab:experiment}
\end{table*}

\subsubsection{Assistants}
In FlexRAG, the RAG assistant encapsulates the entire RAG process. This encapsulation standardizes the interaction between the RAG pipeline and the user, while also streamlining the evaluation of the pipeline. 
Specifically, the RAG assistant should provide a chat interface that accepts user input, generates appropriate responses, and returns both the retrieved results and generated responses, along with relevant metadata.

Furthermore, FlexRAG also incorporates several built-in RAG assistants:
\begin{itemize}
    \item \emph{ModularAssistant}: A modular assistant that can be arbitrarily configured through configuration files.
    \item \emph{OnlineAssistant}: An assistant that leverages online API calls to access RAG services provided by commercial companies.
\end{itemize}

\subsection{Evaluation}
\paragraph{Tasks}
Given the sustained scholarly interest in RAG, researchers have proposed a variety of tasks to assess RAG systems and their individual components.
After a comprehensive review of existing evaluation benchmarks \cite{RAGBenchSurvey,KILT,FlashRAG,MTRAG,MTEB}, we found that these tasks can be categorized into multi-turn dialogue tasks, single-turn question-answering tasks, specialized tasks, and retrieval tasks.
As shown in Figure~\ref{fig:tasks}, these tasks can be further classified into two categories: generative tasks and retrieval-based tasks.
Accordingly, we provide two command-line tools in FlexRAG to evaluate these two types of tasks. To ensure a fairer evaluation process, we have developed pre-configured retrievers for the widely used Wikipedia knowledge base.
These retrievers have been made publicly available on the Hugging Face Hub, providing researchers with a convenient and standardized resource for their evaluations.

\paragraph{Metrics}
FlexRAG supports a variety of evaluation metrics for assessing the performance of RAG systems. These metrics can be broadly categorized into two types: retrieval metrics, and generation metrics.
To ensure the accuracy and reliability of the evaluation results, we employed the widely adopted pytrec\_eval\footnote[13]{https://github.com/cvangysel/pytrec\_eval}, sacreBLEU\footnote[14]{https://github.com/mjpost/sacrebleu}, and Rouge\footnote[15]{https://github.com/pltrdy/rouge} for metric computation.
Additionally, FlexRAG also supports several LLM-as-a-Judge metrics for evaluating the quality of generated responses.

\section{Empirical Study}
To demonstrate the advantages of FlexRAG in research and prototype development, we conducted several experiments on \emph{ModularAssistant}, a highly flexible RAG pipeline within FlexRAG.
We evaluated the performance of the pipeline on three widely used RAG tasks: \emph{Natural Questions}\cite{NQ}, \emph{TriviaQA}\cite{TriviaQA}, and \emph{PopQA}\cite{PopQA}.
We employed the Wikipedia knowledge base provided by \citet{DPR}.
In the experiment, we fixed the other components of the \emph{ModularAssistant} and independently varied its retriever, indexer, re-ranker, and generator to demonstrate the roles played by each component in the RAG task.
Additionally, the number of contexts fed into the generator is fixed at 10. When the reranker is employed, we retrieve 100 contexts from the retriever and employ the reranker to select the top 10 most relevant ones.
We employed the F1 and Exact Match (EM) scores to evaluate the generation quality, and the Success Rate (Succ) to evaluate the retrieval quality.

As shown in Table~\ref{tab:experiment}, the results demonstrate that the choice of retriever, indexer, re-ranker, and generator significantly impacts the overall performance of the RAG system. For more detailed information about the experiments and the experimental findings, please visit our benchmark pages\footnote[16]{\href{https://github.com/ictnlp/FlexRAG/blob/master/benchmarks/singlehop\_qa.md}{https://github.com/ictnlp/FlexRAG/blob/master/\\benchmarks/singlehop\_qa.md}}.

\section{Resource Overhead Analysis}
To further validate the advantages of FlexRAG in terms of system resource efficiency, we evaluated its dense retrieval performance on the \textit{MS\_MARCO Passage Retrieval} \cite{MS_Marco} task using a server equipped with 256 GB of RAM, two Intel Xeon Silver 4214R CPUs, and eight GeForce RTX 3090 GPUs.
FlashRAG, whose architecture is most similar to that of FlexRAG, was selected as the baseline for comparison.
All experiments were conducted under default parameter settings.
his evaluation primarily focuses on the following four system resource metrics:

\begin{itemize}
    \item \textbf{Average Wall-Clock Time}: The average time taken to complete a single retrieval operation. This metric is crucial for assessing the actual latency experienced by users during the retrieval process.
    \item \textbf{Total CPU Time}: The total CPU time consumed during the retrieval process. This metric provides insight into the computational efficiency of the retrieval operation.
    \item \textbf{Average Memory Usage}: The average memory usage during the retrieval process. This metric reflects the memory efficiency of the retrieval operation, which is particularly important for large-scale retrieval tasks.
    \item \textbf{Total Memory Usage}: The total memory usage during the retrieval process.
\end{itemize}

\begin{figure}[t]
  \centering
  \includegraphics[width=1.0\linewidth]{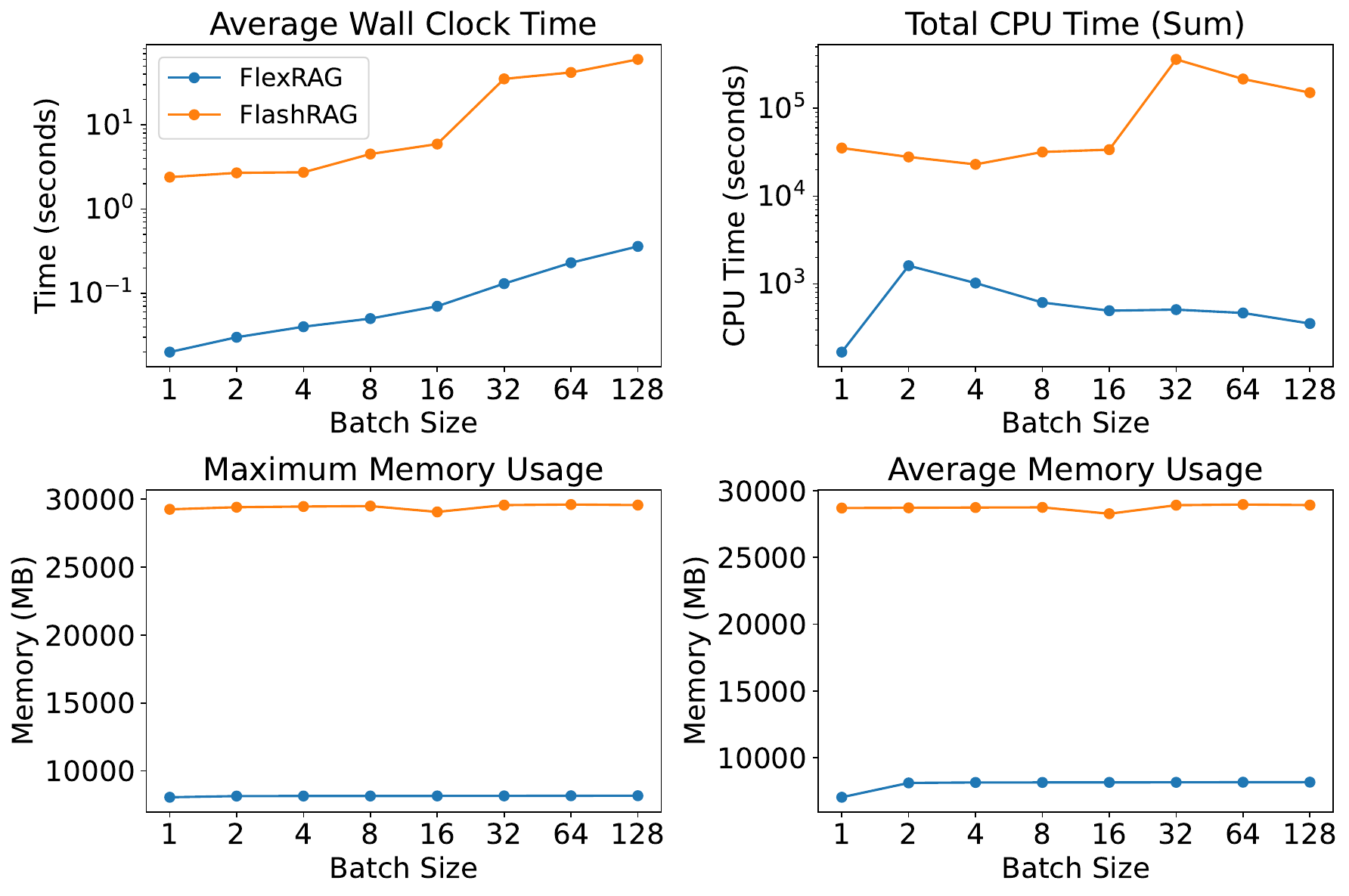}
  \caption{The resource overhead of FlexRAG and FlashRAG under different batch sizes.}
  \label{fig:overhead}
\end{figure}

As shown in Figure~\ref{fig:overhead}, under varying batch sizes, FlexRAG consistently exhibits significantly lower overhead compared to FlashRAG in both average wall-clock time and total CPU time, with performance gaps reaching up to an order of magnitude.
In terms of memory consumption, FlexRAG also demonstrates substantially lower average and peak memory usage, outperforming the baseline by several times.
These results highlight the tangible performance benefits achieved through the incorporation of a memory-mapping mechanism into the system architecture, as well as the optimization of dense index parameters using the ANN-Benchmark toolkit.

Additionally, we observe a general trend wherein system latency increases with larger batch sizes, while total CPU overhead tends to decrease.
A particularly noteworthy case arises when the batch size is set to 1: under this configuration, FlexRAG achieves the lowest computational overhead across all settings.
Further investigation reveals that this outcome stems from the Tokenizer component operating in a single-process mode, thereby avoiding the additional overhead associated with inter-process scheduling.

\section{Conclusion}
In this paper, we introduce FlexRAG, an open-source framework designed to facilitate the rapid reproduction, development, and evaluation of RAG systems.
% In general, FlexRAG significantly reduces the barrier for researchers to construct RAG systems, streamlines collaboration and knowledge sharing, and facilitates a seamless transition from algorithmic research to prototype development through an end-to-end pipeline design.
In general, FlexRAG significantly reduces the barrier to building RAG systems, streamlines collaboration, and enables a seamless transition from research to prototyping with an integrated pipeline design.

\bibliography{anthology,custom}

\appendix

\newpage

% \usepackage{graphicx}
\begin{table*}[t]
\resizebox{\textwidth}{!}{%
\begin{tabular}{llllllll}
\toprule
Frameworks                                      & Web Access & Multimodal & Preprocess & Evaluation & Training Scripts & Research Oriented & Still Maintain \\
\midrule
LangChain\footnotemark[1]               & \ding{51}  & \ding{51}   & \ding{51}  & \ding{51}  & \ding{51}        & \ding{55}         & \ding{51}      \\
LlamaIndex\footnotemark[2]              & \ding{51}  & \ding{51}   & \ding{51}  & \ding{51}  & \ding{51}        & \ding{55}         & \ding{51}      \\
FlashRAG\cite{FlashRAG}                 & \ding{55}  & \ding{51}   & \ding{55}  & \ding{51}  & \ding{55}        & \ding{51}         & \ding{51}      \\
RAGLab\cite{RAGLab}                     & \ding{55}  & \ding{55}   & \ding{55}  & \ding{51}  & \ding{51}        & \ding{51}         & \ding{55}      \\
AutoRAG\cite{AutoRAG}                   & \ding{55}  & \ding{55}   & \ding{51}  & \ding{51}  & \ding{55}        & \ding{51}         & \ding{51}      \\
AutoRAG-HP\cite{AutoRAG_HP}             & \ding{55}  & \ding{55}   & -          & \ding{51}  & \ding{55}        & \ding{51}         & -              \\
RaLLe\cite{Ralle}                       & \ding{55}  & \ding{55}   & \ding{55}  & \ding{51}  & \ding{55}        & \ding{51}         & \ding{55}      \\
LocalRQA\cite{LocalRQA}                 & \ding{55}  & \ding{55}   & \ding{51}  & \ding{51}  & \ding{51}        & \ding{51}         & \ding{55}      \\
EasyRAG\cite{EasyRAG}                   & \ding{51}  & \ding{55}   & \ding{55}  & \ding{55}  & \ding{55}        & \ding{51}         & \ding{55}      \\
UltraRAG\footnotemark[3]                & \ding{55}  & \ding{51}   & \ding{55}  & \ding{51}  & \ding{51}        & \ding{51}         & \ding{51}      \\
\textbf{FlexRAG (Ours)}                 & \ding{51}  & \ding{51}   & \ding{51}  & \ding{51}  & \ding{55}        & \ding{51}         & \ding{51}      \\
\bottomrule
\end{tabular}%
}
\caption{Comparison with existing retrieval-augmented generation frameworks. We evaluate each framework based on the following criteria: (1) support for internet access, (2) multimodal RAG capabilities, (3) inclusion of preprocessing modules, (4) availability of evaluation modules, (5) provision of training scripts, (6) research-oriented design, and (7) active maintenance status (defined as having commits within the last three months). "-" indicates the framework is not currently public available.}
\label{tab:comparision}
\end{table*}

\footnotetext[1]{https://www.langchain.com/}
\footnotetext[2]{https://www.llamaindex.ai/}
\footnotetext[3]{{https://github.com/OpenBMB/UltraRAG}}

\section{Comparison with Existing RAG Frameworks}
\label{sec:appendix}
To further illustrate the uniqueness of FlexRAG, we conducted a comparative analysis of a wide range of related works. The results of this comparison are summarized in Table~\ref{tab:comparision}.
As a heavyweight framework, LangChain and LlamaIndex offer the most comprehensive functionalities.
However, the research-oriented design brings FlexRAG distinct advantages in algorithm reproducibility and knowledge sharing.
At the same time, its lightweight architecture ensures a smoother learning curve, making it more accessible to researchers and developers alike.

Among lightweight frameworks, FlashRAG has made notable contributions to the reproducibility of existing researches. Beyond this, FlexRAG offers a more extensive set of fundamental components, supports web access, integrates seamlessly with Hugging Face, and features a well-structured preprocessing module.
UltraRAG incorporates numerous cutting-edge techniques. In contrast, the modular architecture of FlexRAG allowing researchers to efficiently extend and customize it to meet their evolving needs.
Meanwhile, AutoRAG and AutoRAG-HP focus primarily on automated hyperparameter tuning, while several other frameworks in this category have been discontinued.

\end{document}